\documentclass{article}
\usepackage{titlesec}
\usepackage{graphicx}
\usepackage{arxiv}
\usepackage[utf8]{inputenc}
\usepackage{setspace}
\usepackage{geometry}
\usepackage{indentfirst}
\usepackage{amsmath}
\usepackage{graphicx}
\usepackage{xcolor}
\definecolor{verylightgray}{gray}{0.9}

\usepackage{natbib}
\usepackage[T1]{fontenc}    
\usepackage{hyperref}       
\usepackage{url}            
\usepackage{booktabs}       
\usepackage{amsfonts}       
\usepackage{nicefrac}       
\usepackage{microtype}      
\usepackage{lipsum}		
\usepackage{doi}

\title{Structural Equation--VAE: Disentangled Latent Representations for Tabular Data}
\date{}  

\author{ 
	\href{https://orcid.org/0000-0002-0883-4574}{\includegraphics[scale=0.06]{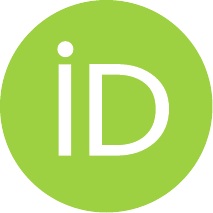}\hspace{1mm}Ruiyu Zhang}\\
	Department of Politics and Public Administration\\
    The University of Hong Kong\\
	ruiyuzh@connect.hku.hk\\
	\And
        \href{ }
	{\hspace{1mm}Ce Zhao} \\
	School of Computer Science\\
    Carnegie Mellon University\\
        alcor\_zhao@outlook.com\\
        \And
	\href{https://orcid.org/0009-0005-7399-109X}{\includegraphics[scale=0.06]{orcid.pdf}\hspace{1mm}Xin Zhao}\\
	Department of Applied Social Sciences\\
    The Hong Kong Polytechnic University\\
        \And
	\href{https://orcid.org/0000-0002-0275-117X}{\includegraphics[scale=0.06]{orcid.pdf}\hspace{1mm}Lin Nie}\\
	Department of Applied Social Sciences\\
    The Hong Kong Polytechnic University\\
        \And
	\href{https://orcid.org/0000-0002-2303-0996}{\includegraphics[scale=0.06]{orcid.pdf}\hspace{1mm}Wai-Fung Lam}\\
	Department of Politics and Public Administration\\
    The University of Hong Kong
}

\renewcommand{\headeright}{ }
\renewcommand{\undertitle}{ }
\renewcommand{\shorttitle}{Structural Equation–VAE}

\hypersetup{
    pdftitle={Structural Equation–VAE Zhang et.al},
    pdfauthor={Ruiyu Zhang, Ce Zhao, Xin Zhao, Lin Nie, Wai-Fung Lam},
    pdfkeywords={Disentangled Representation Learning, Tabular Data, Variational Autoencoder, Structural Equation Modeling, Latent Variable Modeling},
}

\begin{document}
\maketitle
\footnotetext[1]{The SE-VAE Python package can be installed via: \colorbox{verylightgray}{\texttt{pip install sevae}}. We provide SE-VAE as an off-the-shelf, easy-to-use package so that researchers can directly apply it in their work. For documentation and tutorials, visit: \url{https://pypi.org/project/sevae/}.}

\begin{abstract}
    Learning interpretable latent representations from tabular data remains a challenge in deep generative modeling. We introduce SE-VAE (Structural Equation–Variational Autoencoder), a novel architecture that embeds measurement structure directly into the design of a variational autoencoder. Inspired by structural equation modeling, SE-VAE aligns latent subspaces with known indicator groupings and introduces a global nuisance latent to isolate construct-specific confounding variation. This modular architecture enables disentanglement through design rather than through statistical regularizers alone. We evaluate SE-VAE on a suite of simulated tabular datasets and benchmark its performance against a series of leading baselines using standard disentanglement metrics. SE-VAE consistently outperforms alternatives in factor recovery, interpretability, and robustness to nuisance variation. Ablation results reveal that architectural structure, rather than regularization strength, is the key driver of performance. SE-VAE offers a principled framework for white-box generative modeling in scientific and social domains where latent constructs are theory-driven and measurement validity is essential.
\end{abstract}


\section{Introduction}

Learning compact, interpretable representations from structured, tabular data remains a central challenge in unsupervised machine learning\citep{bengio2013representation}. While representation learning has made substantial advances in visual and sequential domains, most deep generative models—particularly Variational Autoencoders (VAEs)—struggle to produce disentangled and interpretable latent variables when applied to tabular datasets. \citep{kingma2013auto, higgins2017beta, kim2018disentangling, locatello2019challenging} This is problematic in scientific domains where tabular data are common, and where interpretability, construct validity, and transparency are not optional but foundational. \citep{bollen1989structural, tabri2012principles} Unlike images, where factors of variation are often spatial and continuous, tabular datasets typically encode measurements of latent constructs—such as attitudes, traits, biological pathways, or experimental conditions—via structured groups of indicators. \citep{bollen1989structural} In these settings, standard VAEs offer little control over what each latent dimension encodes, resulting in representations that are opaque, entangled, and vulnerable to nuisance variation.

To address these limitations, we propose a new architecture: the \textbf{SE-VAE} (Structural Equation–Variational Autoencoder), a white-box generative model tailored to tabular data with known indicator–construct structure. SE-VAE is designed to be interpretable by construction. It imposes structural constraints on the latent space that reflect prior knowledge about how observed variables map to latent constructs. Specifically, each group of observed indicators is assigned a dedicated latent subspace, enabling modular and factor-aligned encoding. In addition, SE-VAE introduces explicit nuisance latent variables to capture confounding structure that spans across observed variables, thereby isolating the latent constructs of interest. Rather than focusing on enforcing disentanglement through implicit statistical penalties alone (as in $\beta$-VAE or FactorVAE), \citep{higgins2017beta, kim2018disentangling} SE-VAE incorporates \emph{measurement structure} as an architectural inductive bias—resulting in more stable, transparent, and purpose-aligned representations.

SE-VAE draws inspiration from Structural Equation Modeling (SEM), a classical approach widely used in both the natural and social sciences to relate observed variables to latent constructs via pre-specified loading patterns. \citep{bollen1989structural, tabri2012principles} While SEM offers strong interpretability, it is constrained by linearity, fixed functional forms, and limited scalability. SE-VAE relaxes these constraints through the flexibility of neural architectures and amortized inference, while preserving SEM's core advantage: the ability to align latent dimensions with theory-driven constructs. This integration allows SE-VAE to learn meaningful representations in domains where measurement theory matters—ranging from psychometrics and genomics to economics and experimental science.

We evaluate SE-VAE on a battery of simulations designed to reflect real-world tabular complexity: multiple latent factors, structured item groupings, nonlinear shared confounding, and varying sample sizes. Disentanglement quality is assessed using four widely adopted metrics—Mutual Information Gap (MIG), Disentanglement–Completeness–Informativeness (DCI), Separated Attribute Predictability (SAP), and Permutation Alignment\citep{chen2018isolating, eastwood2018framework, kumar2018variational, locatello2019challenging, kuhn1955hungarian}—and benchmarked against $\beta$-VAE and FactorVAE. Our results show that SE-VAE consistently outperforms baselines in recovering ground-truth latent factors, maintaining robustness to confounding variation, and offering interpretable representations that align with known measurement structures. Ablation experiments further demonstrate the distinct contributions of factor-aligned subspaces and nuisance disentanglement.

By embedding measurement structure directly into the generative model, SE-VAE advances the field of white-box representation learning for tabular data. It offers a scalable, theoretically grounded alternative to black-box approaches, and opens new directions for applying generative modeling in domains where meaning, not just reconstruction, is the objective.

\section{Related Work}

Unsupervised representation learning has sought to extract disentangled latent factors from high-dimensional data, enabling interpretability and modularity. Early efforts such as the $\beta$-VAE introduced a simple yet powerful modification to the standard VAE objective, increasing the KL-divergence penalty to promote disentanglement between latent dimensions \cite{higgins2017beta}. Extensions like FactorVAE \cite{kim2018disentangling} and DIP-VAE \cite{kumar2018variational} refined this approach by directly penalizing dependencies in the aggregated posterior. A recent comprehensive review consolidates the growing body of disentanglement methods,\citep{wang2024disentangled} emphasizing the diversity of inductive biases—architectural, statistical, and contrastive—that underpin successful representation learning. However, Locatello et al. \cite{locatello2019challenging} demonstrated that, without supervision or inductive biases, disentanglement is fundamentally unidentifiable—no unsupervised objective alone can guarantee recovery of the true latent factors, especially in the absence of structural constraints.

To address this limitation, a parallel literature has explored identifiability via auxiliary information. iVAE \cite{khemakhem2020variational} shows that if latent variables are conditionally independent given observed covariates, identifiability is possible under mild assumptions. These models often rely on conditional priors, side information, or domain supervision to constrain the latent space. Nonetheless, such approaches typically assume continuous covariates or rely on strong conditional independence assumptions that are not always practical in tabular domains with latent measurement structures. Recent advances in neural architectures for tabular data, highlight the unique challenges and design considerations when applying deep learning to structured inputs. \citep{gorishniy2021revisiting} However, these models primarily focus on predictive accuracy and offer limited support for disentangled or interpretable latent representations.

Another related strand involves nuisance-invariant representation learning. Variational Fair Autoencoders (VFAE) \cite{louizos2015variational} and adversarial learning frameworks \cite{edwards2015censoring} attempt to remove unwanted sources of variation—such as sensitive attributes—by encouraging the latent representation to be invariant to known confounders. While effective in fairness applications, these methods often rely on adversarial optimization or explicit domain labels and do not generalize easily to tabular settings where nuisance structure is unknown or nonlinear.

A growing body of work on structured generative models has aimed to incorporate inductive biases directly into the model architecture. Semi-supervised VAEs \cite{kingma2014semi} and Structured VAEs (S-VAE) \cite{johnson2016composing} have used graphical model formulations to blend probabilistic structure with deep learning flexibility. These models offer strong foundations for hybrid generative inference but are not typically designed for measurement-aligned encoding or disentanglement in tabular contexts.

Finally, disciplines such as psychometrics, economics, and the natural sciences have long used latent variable models to encode constructs like intelligence, ability, or unobserved shocks. Structural Equation Modeling (SEM) \cite{bollen1989structural, tabri2012principles} and latent factor analysis encode prior knowledge through loading structures, making latent dimensions interpretable by design. Yet, these classical models assume linearity and Gaussianity, and are not readily scalable to complex, nonlinear domains. Our proposed SE-VAE integrates the architectural transparency of SEM with the expressiveness of deep generative models, providing an interpretable, scalable solution for representation learning in tabular scientific data.

\section{The SE-VAE Model}

The Structural Equation–Variational Autoencoder (SE-VAE) is designed to embed measurement structure directly into the architecture of deep generative models. Its core motivation is to enable interpretable and disentangled latent representations in tabular datasets where each set of observed variables (i.e., indicators) is theorized to reflect an underlying latent construct. This is especially important in scientific applications where the latent dimensions are not abstract or arbitrary, but instead correspond to meaningful and measurable factors.

\begin{figure}[htbp]
    \centering
    \includegraphics[width=0.95\textwidth]{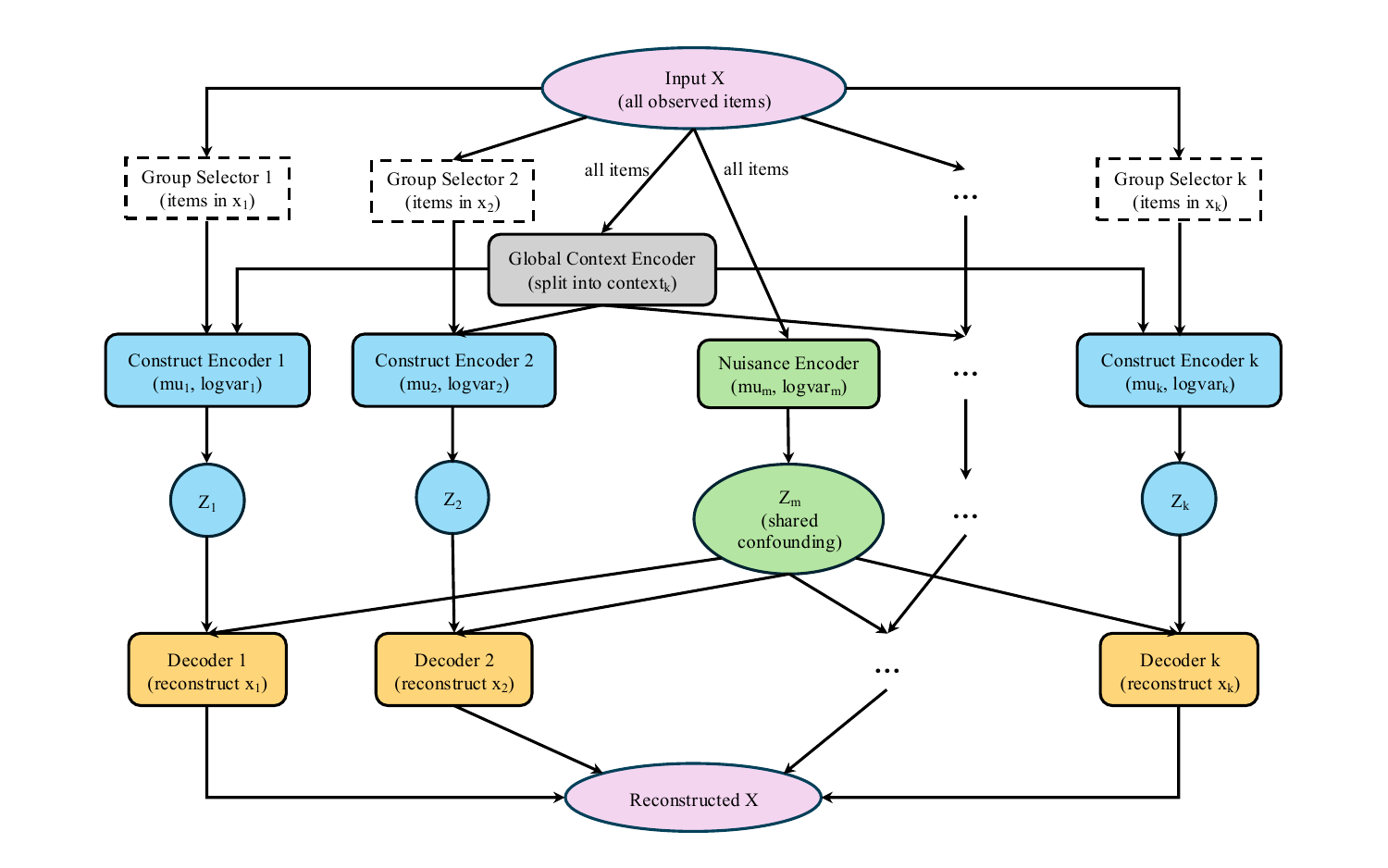}
    \caption{SE-VAE architecture: each indicator group $x_k$ is encoded by a construct-specific encoder, enhanced with context from a shared Global Context Encoder. A Nuisance Encoder extracts $z_m$ to absorb cross-group variation. Each decoder reconstructs $x_k$ from $[z_k, z_m]$.}
    \label{fig:sevae_architecture}
\end{figure}

At the heart of SE-VAE is a partitioned architecture that mirrors the indicator–construct structure commonly found in measurement theory. The encoder is divided into multiple parallel sub-encoders, each dedicated to a specific group of observed indicators. These sub-encoders extract distinct latent subspaces $z_k$, where $k$ indexes the latent constructs. Each $z_k$ is derived from a concatenation of its corresponding group of input variables $x_k$ and a shared context vector $c_k = f_{\text{context}}(X)$, where $f_{\text{context}}$ is a global context encoder that processes the full input $X$ to produce low-dimensional representations of shared patterns.
 This context encoder processes the full input vector $X$ and provides each sub-encoder with a low-dimensional representation of global patterns. The inclusion of this global context mechanism enables each construct encoder to flexibly adjust for weak inter-indicator correlations while maintaining its primary responsibility for local structure. This fusion of local and contextual information yields variational parameters $(\mu_k, \log \sigma_k^2)$ that define the posterior distribution $q(z_k \mid x_k, X)$.

To isolate variation that is not construct-specific—such as measurement artifacts, data quality heterogeneity, or systematic confounding—the SE-VAE introduces an additional shared nuisance latent $z_m$. Unlike the partitioned sub-latents, $z_m$ is inferred from the entire input and represents global patterns or noise that span across indicator groups. This design ensures that the sub-latents $z_k$ capture construct-specific variation, while $z_m$ absorbs extraneous influences that are irrelevant or disruptive to the interpretation of individual factors.

The decoder in SE-VAE is similarly modular. Each output group $x_k$ is reconstructed solely from its corresponding sub-latent $z_k$ and the shared nuisance latent $z_m$. This architectural constraint enforces local generative responsibility and supports transparency: the reconstruction of each indicator group is driven only by its assigned latent dimension and a global residual component. To ensure that $z_m$ does not encode construct-specific information, SE-VAE incorporates an adversarial leakage loss. For each group $x_k$, an auxiliary decoder attempts to reconstruct $x_k$ using only $z_m$. The main decoder reconstructs $x_k$ from $[z_k, z_m]$, and the reconstruction loss from the auxiliary decoder is used to penalize leakage from $z_m$.This adversarial training component encourages a cleaner separation between nuisance variation and construct-specific structure, particularly when indicator groups exhibit overlapping variance due to latent confounders or shared measurement noise. By penalizing leakage, SE-VAE ensures that $z_m$ remains focused on global nuisance signals and does not interfere with the interpretability of individual latent constructs.

Formally, the adversarial leakage loss for a mini-batch of size $B$ with $K$ indicator groups, each of dimensionality $J$, is defined as:
\[
\mathcal{L}_{\text{adv}} = \frac{1}{K} \sum_{k=1}^{K} \left[ 
    \underbrace{\mathrm{MSE}(x_k, \hat{x}_k)}_{\text{Main decoder: } [z_k, z_m]}
    + \lambda_{\text{adv}} \cdot \underbrace{\mathrm{MSE}(x_k, \hat{x}_k^{\text{adv}})}_{\text{Adversarial decoder: } z_m \text{ only}}
\right]
\]
where $\hat{x}_k$ is the reconstruction from the main decoder using $[z_k, z_m]$, and $\hat{x}_k^{\text{adv}}$ is the adversarial reconstruction using only $z_m$. The hyperparameter $\lambda_{\text{adv}}$ controls the strength of the adversarial penalty.

\vspace{2em}

The generative model formalizes this structure as follows. For each indicator group $x_k$, we assume a conditional distribution of the form $p(x_k \mid z_k, z_m)$. The latent variables $z_k$ and $z_m$ are drawn from standard Gaussian priors: $z_k, z_m \sim \mathcal{N}(0, I)$. This formulation supports amortized variational inference while maintaining interpretable latent structure.

To train the model, SE-VAE maximizes an augmented evidence lower bound (ELBO) that incorporates both standard reconstruction and KL divergence terms, as well as additional losses that promote disentanglement and interpretability. Specifically, a Total Correlation (TC) penalty encourages independence across latent dimensions, and an orthogonality constraint across the $z_k$ spaces further enforces factor separation. Together, these objectives ensure that the latent space remains structured, sparse, and aligned with the underlying theoretical constructs.

SE-VAE thus combines architectural modularity with explicit nuisance separation, enabling white-box generative modeling in settings where latent constructs are grounded in domain knowledge and measurement theory. This design allows researchers to use generative models not only for data compression, but for theory-driven inference and discovery in complex tabular domains. 

SE-VAE potentially addresses longstanding challenges faced by scientific and engineering disciplines, where researchers have struggled to infer latent structure under conditions of entangled noise and partial observability. In neuroscience, efforts to isolate task-relevant neural representations are often confounded by shared nuisance signals such as scanner drift, motion artifacts, or arousal-related fluctuations. SE-VAE offers a potential path forward by explicitly modeling nuisance variation in a dedicated latent space, thereby facilitating cleaner recovery of construct-specific signals. In control systems engineering, the simultaneous estimation of interpretable system states and unstructured environmental disturbances has proven difficult to formalize without compromising model identifiability. The modular architecture of SE-VAE may help to disentangle these sources by assigning local and global generative responsibilities. In the social sciences, persistent issues such as common method bias continue to threaten construct validity in observational and survey-based research. By incorporating structured measurement assumptions and a principled nuisance separation mechanism, SE-VAE holds promise for psychometric applications where theoretical constructs must be inferred with transparency and robustness. Across these domains, SE-VAE provides a flexible framework for theory-aligned generative modeling under complex data conditions.

\subsection{Theoretical Analysis}

A key objective of SE-VAE is to achieve modular latent factorization that aligns with a known item–construct structure. This is made possible by explicitly partitioning the encoder and decoder pathways, incorporating global contextual information, and enforcing separation between the latent subspaces assigned to different indicator groups. In this section, we theoretically motivate the design of SE-VAE by examining its ability to (1) recover latent subspace factorization under grouped indicators, (2) separate nuisance variation, and (3) introduce a strong inductive bias for identifiability in representation learning.

\paragraph{Latent Subspace Factorization.}  
Consider a dataset where observed variables $\mathbf{x} = [x_1, \dots, x_K]$ are grouped into $K$ mutually exclusive indicator sets, each generated by a corresponding latent construct $z_k$. In standard VAEs, the latent space is entangled—each latent dimension may absorb signals from multiple constructs, leading to poor interpretability. SE-VAE circumvents this by enforcing subspace modularity: each encoder maps $x_k$ and a global context vector $c_k = f_{\text{context}}(X)$ to a latent variable $z_k$, and each decoder reconstructs $x_k$ only from $[z_k, z_m]$.

This architectural separation enables SE-VAE to recover a block-structured latent representation, where each $z_k$ captures variance local to its associated indicator group. Moreover, the addition of a global nuisance latent $z_m$ allows the model to absorb shared noise or cross-cutting artifacts that would otherwise corrupt subspace purity. An adversarial loss ensures that $z_m$ does not encode construct-specific variation, thereby preserving the interpretability of each $z_k$.

Formally, let $q(z_k \mid x_k)$ denote the variational posterior for the $k$-th construct latent, and $q(z_j \mid x_j)$ for another group $j \neq k$. The architecture enforces statistical independence between $z_k$ and $z_j$ by design. If we further impose an orthogonality constraint $\mathbb{E}[z_k^\top z_j] = 0$ during training, the resulting latent subspaces become decorrelated. When the data-generating process is modular—i.e., $x_k \perp x_j$ conditioned on their respective $z_k, z_j$—this design recovers the ground-truth modular structure. Empirically, we find that orthogonal regularization improves disentanglement without sacrificing reconstruction fidelity.

\paragraph{Inductive Bias and Identifiability.}  
Locatello et al. (2019) provide a fundamental impossibility result: disentangled representations cannot be identified in an unsupervised setting without inductive bias.\citep{locatello2019challenging} Scholars further demonstrate that contrastive learning, when carefully designed, can invert the data-generating process, underscoring how architectural and loss-based inductive biases may unlock identifiability.\citep{zimmermann2021contrastive} In other words, without guidance from supervision or architectural assumptions, VAEs may learn latent spaces that fit the data but fail to align with meaningful or interpretable generative factors—making true disentanglement impossible. In response, many prior works have introduced statistical regularizers (e.g., $\beta$-VAE, FactorVAE, DIP-VAE) to bias the solution space toward disentanglement. However, these regularizers operate in a black-box fashion and often fail in tabular settings where statistical redundancy is low.

SE-VAE introduces an alternative form of inductive bias—one grounded in architectural design that reflects the known structure of measurement models. By constraining the input–output pathways such that each latent $z_k$ is derived solely from and responsible for a specific group of indicators $x_k$, SE-VAE embeds domain knowledge about indicator–construct relationships directly into the model. 

This structural prior can be formalized through a partially factorized posterior, in which each construct-specific latent variable depends only on its corresponding observed group, while the share confounding latent is inferred from the full input:

This structural prior can be formalized through a partially factorized posterior, in which each construct-specific latent variable depends on both its corresponding observed group and a global context vector $c_k$, while the shared nuisance latent is inferred directly from the full input:

\begin{equation}
q(\mathbf{z} \mid \mathbf{x}) = q(z_m \mid \mathbf{x}) \cdot \prod_{k=1}^{K} q(z_k \mid x_k, c_k)
\label{eq:posterior_factorization}
\end{equation}

Meanwhile, the decoder preserves modular generative paths by reconstructing each observed group using only the corresponding sub-latent and the shared confounding:

\begin{equation}
p(\mathbf{x} \mid \mathbf{z}) = \prod_{k=1}^{K} p(x_k \mid z_k, z_m)
\label{eq:decoder_modularity}
\end{equation}

This item group exclusivity acts as a structural prior, narrowing the hypothesis space and enabling identifiability up to isomorphism under the assumed factor structure. In other words, SE-VAE makes the disentanglement problem solvable not by statistical tricks, but by aligning the generative process with human-understood measurement logic.

This approach draws parallels to identifiability in structural equation modeling, where latent constructs are defined via unique loading patterns over observed variables. While classical SEM assumes linearity and fixed forms, SE-VAE generalizes this logic to nonlinear generative modeling, while retaining interpretability. By combining encoder partitioning, global context integration, decoder locality, adversarial leakage suppression, and orthogonal regularization, SE-VAE creates the conditions under which the model’s latent dimensions reflect distinct, meaningful sources of variation.

\section{Experimental Setup}

To evaluate SE-VAE, we design a controlled simulation benchmark where the ground-truth latent structure is explicitly aligned with the model’s measurement architecture. This allows for a fair and interpretable comparison with baseline models and ablation variants.

\subsection{Synthetic Data Generation}

We simulate a tabular dataset with $N=50{,}000$ samples and $K \in \{4, 6, 8, 10, 12\}$ latent constructs, each associated with $J=6$ observed indicators. Each true latent factor is independently sampled from a standard Gaussian, and each indicator is generated through a nonlinear transformation of its corresponding latent, with additional noise, cross-loadings (15\% of items), and two-way interactions. A shared confounding is also sampled and injected into each item using nonlinear effects (including sine and quadratic terms), scaled by item-specific shared confounding loadings. This design mimics realistic measurement challenges such as heteroskedasticity, nonlinearities, and confounding.

\subsection{Baselines and Variants}

We compare SE-VAE against five widely used disentanglement baselines: (1) the original VAE \citep{kingma2013auto}, which serves as a non-regularized benchmark; (2) the Variational Fair Autoencoder (VFAE) \citep{louizos2015variational}, which incorporates a fairness-oriented objective to remove sensitive attribute information from latent variables; (3) $\beta$-VAE \citep{higgins2017beta}, which increases the KL divergence weight to encourage disentanglement; (4) FactorVAE \citep{kim2018disentangling}, which introduces an explicit penalty on total correlation among latent variables; and (5) DIP-VAE \citep{kumar2018variational}, which aligns the aggregated posterior with a factorial prior using moment-matching regularization; and (6) $\beta$-TCVAE \citep{chen2018isolating}, which decomposes the ELBO and penalizes the total correlation term with a specific weight to improve disentanglement. For each baseline, we experiment with multiple levels of their respective regularization strengths to ensure a fair and comprehensive evaluation. All models share the same encoder–decoder architecture, consisting of multilayer perceptrons with two hidden layers and ReLU activations. The number of latent dimensions is fixed to match the number of ground-truth factors in the simulated datasets. We train on data sizes ranging from 2,000 to 50,000 samples to evaluate robustness to sample size.

\subsection{Evaluation Metrics}

To assess the quality of learned representations, we employ four complementary metrics that quantify different aspects of disentanglement. First, the \textit{Mutual Information Gap} (MIG) \citep{chen2018isolating} measures the extent to which each true generative factor is uniquely captured by one latent dimension. Specifically, it computes the difference in mutual information between the most and second-most informative latents for each factor, normalized by the entropy of the factor. Second, the \textit{Disentanglement–Completeness–Informativeness} (DCI) framework \citep{eastwood2018framework} evaluates disentanglement as the concentration of predictive power for each factor within a single latent, completeness as the degree to which each latent is specific to a particular factor, and informativeness as the average $R^2$ in predicting factors from latents. Third, the \textit{Separated Attribute Predictability} (SAP) score \citep{kumar2018variational} compares the best-performing single latent regressor against a joint model across all latents, providing a robustness check for single-latent identifiability. Finally, we compute a \textit{Permutation Alignment Accuracy}, which uses the Hungarian algorithm to optimally match latent dimensions to ground-truth factors based on Pearson correlation, reporting the average absolute correlation of aligned pairs.\citep{locatello2019challenging, kuhn1955hungarian} All evaluations are performed on the mean of the approximate posterior (i.e., $\mu$), and results are averaged across factors to ensure comparability across models and datasets.

\subsection{Ablation Study}

To understand the individual contribution of each architectural and regularization component in SE-VAE, we conduct a comprehensive ablation study. This involves systematically varying the model’s three key loss components: the KL divergence penalty ($\beta$), the total correlation penalty ($\gamma$), and the orthogonality constraint ($\alpha$). Each component is toggled between an active and inactive state, forming a grid of combinations that isolate the presence or absence of each regularizer. In addition, we evaluate the effect of KL annealing by including or omitting it during training, which gradually introduces the KL term to prevent premature posterior collapse. All SE-VAE variants are trained under identical conditions and evaluated using a suite of disentanglement metrics. This ablation strategy allows us to identify which regularization mechanisms and architectural constraints are most critical for producing structured, modular, and interpretable latent representations.

\section{Results}

\subsection{Disentanglement Performance Across Sample Sizes}

\begin{figure}[htbp]
    \centering
    \includegraphics[width=0.83\textwidth]{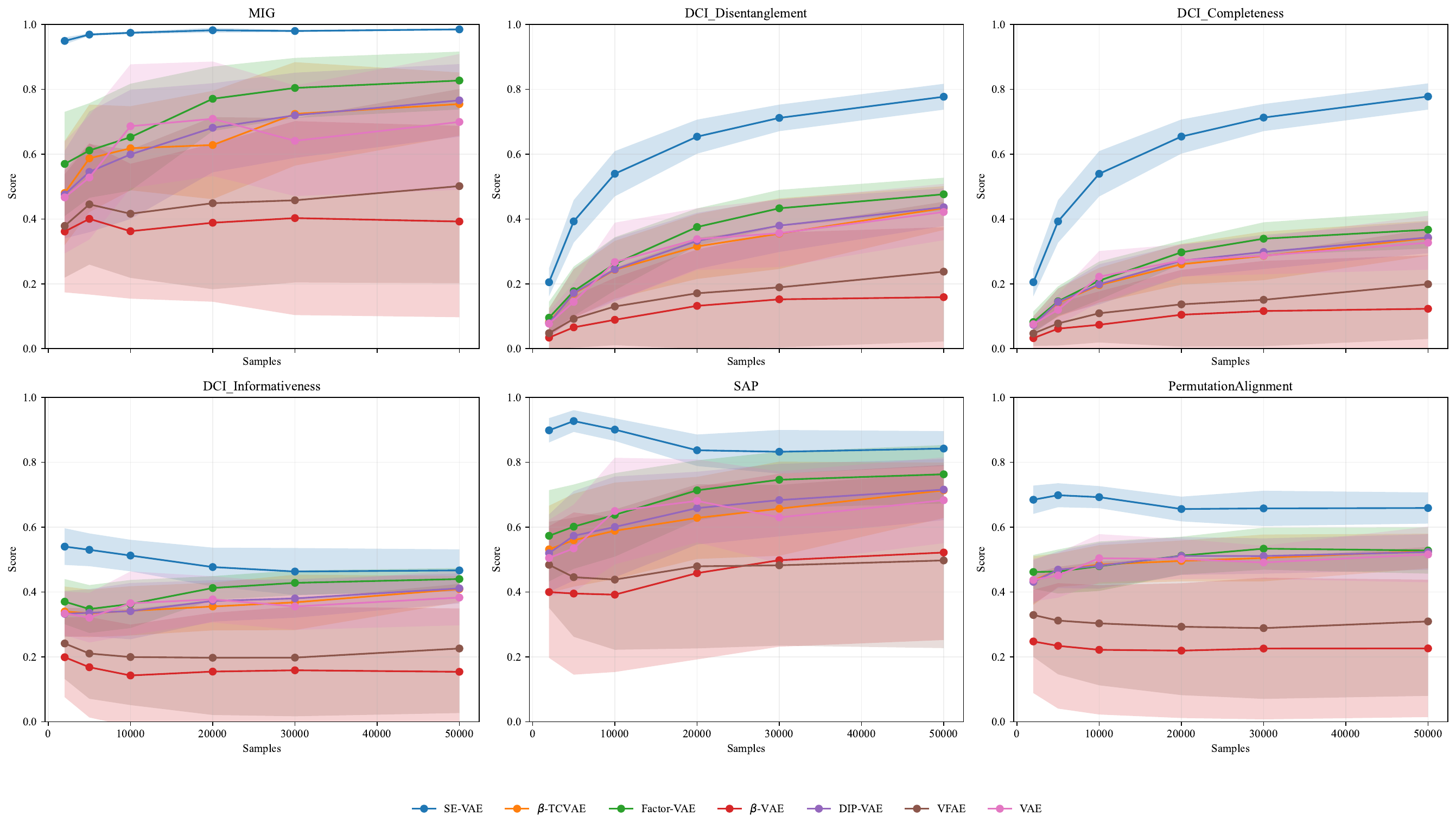}
    \caption{Disentanglement performance across sample sizes (2,000 to 50,000). SE-VAE consistently outperforms baselines across six metrics.}
    \label{fig:disentanglement_results}
\end{figure}

Figure~\ref{fig:disentanglement_results} summarizes the disentanglement performance of SE-VAE compared to six widely used baselines—VAE, VFAE, $\beta$-VAE, FactorVAE, DIP-VAE, and $\beta$-TCVAE—across a range of training sample sizes. Performance is evaluated using six established metrics: DCI Disentanglement, DCI Completeness, DCI Informativeness, MIG, SAP, and Permutation Alignment. Each metric captures a distinct aspect of disentangled representation: DCI Disentanglement reflects axis alignment of latent dimensions with ground-truth factors, Completeness measures how concentrated each factor is in a single latent unit, and Informativeness assesses overall predictive utility. MIG and SAP evaluate mutual information gap and classifier separation, respectively, while Permutation Alignment quantifies direct correspondence to ground-truth factors.

Across all metrics and data sizes, SE-VAE consistently outperforms the baselines by a substantial margin. Gains are most pronounced for \textbf{DCI Completeness} and \textbf{DCI Disentanglement}, where SE-VAE demonstrates rapid improvement with increasing data and reaches near-optimal scores ($>0.75$) with only 30,000–50,000 samples. Notably, SE-VAE also maintains strong \textbf{MIG} and \textbf{SAP} scores, indicating both high mutual information with ground-truth factors and clear factor predictability. These improvements are consistent across runs, as reflected in the relatively narrow standard deviation bands.

While FactorVAE, DIP-VAE and $\beta$-TCVAE exhibit moderate performance improvements with larger datasets, their progress plateaus earlier and fails to match SE-VAE’s interpretability. $\beta$-VAE and VFAE show unstable gains, often peaking at moderate data sizes and declining as dimensionality increases or nuisance interference rises. The original VAE baseline underperforms throughout, reaffirming the need for structured inductive bias in disentanglement tasks.

The stability and scalability of SE-VAE suggest that its modular architecture—combining grouped encoders, global context, orthogonal subspaces, and adversarial nuisance control—offers superior inductive guidance in low-redundancy, tabular settings. By embedding theoretical measurement structure into the model design, SE-VAE more effectively partitions meaningful variation from noise. These results reinforce the value of incorporating structural priors in disentangled representation learning, especially in domains where theoretical construct boundaries are known or inferable. As such, SE-VAE holds practical relevance for scientific inference tasks ranging from bioinformatics to psychometrics.

\subsection{Ablation Analysis of SE-VAE Components}

To evaluate the contribution of SE-VAE's individual components, we conduct an ablation study in which four elements—KL divergence penalty ($\beta$), total correlation penalty ($\gamma$), orthogonality constraint ($\alpha$), and KL annealing—are each toggled between inactive and active states. Each variant is trained under identical conditions and evaluated on six disentanglement metrics across a range of data sizes.

The results show that activating these regularization components leads to only modest performance changes. For example, activating the total correlation penalty increases SAP by approximately 0.24 at 5,000 samples, but this effect diminishes as the data scale increases. Similarly, the orthogonality constraint yields mild improvements in DCI metrics and permutation alignment at smaller sample sizes, but these effects taper off as data grows. The KL divergence penalty and KL annealing exhibit minimal or negligible impact across all settings, with metric changes largely within the range of random variation.

These findings suggest that SE-VAE's strong disentanglement performance does not hinge on any single regularizer. Rather, the improvements stem from its architectural structure: the partitioned encoder-decoder design, the use of grouped latent variables, and the separation of nuisance variation through a dedicated latent space. By enforcing modular generative responsibility and aligning the latent space with measurement theory, SE-VAE achieves disentanglement through design rather than post hoc penalization.

In contrast to prior models that rely heavily on tuning regularization strength, SE-VAE demonstrates stability and robustness even with minimal reliance on auxiliary losses. This reinforces the central claim of the model: that embedding theoretical structure into the model architecture itself offers a more principled and reliable path toward interpretable representation learning, especially in domains where construct boundaries are meaningful and known in advance.

\begin{table}[htbp]
\centering
\caption{Ablation analysis of SE-VAE components. Each row shows the change in disentanglement metrics when activating a specific regularizer or architectural constraint, holding all else constant. Positive values indicate performance gains.}
\label{tab:ablation}
\tiny
\begin{tabular}{lllllllll}
\hline
 Ablation     & Samples   & inactive→active   & MIG            & DCI-D          & DCI-C          & DCI-I          & SAP            & PermAlign      \\
\hline
 	KL penalty ($\beta$)    & 2000      & (inactive→active) & 0.002 ± 0.009  & 0.001 ± 0.020  & 0.002 ± 0.020  & -0.000 ± 0.029 & -0.005 ± 0.049 & 0.001 ± 0.023  \\
              & 5000      &                   & 0.004 ± 0.009  & -0.001 ± 0.031 & 0.000 ± 0.031  & -0.001 ± 0.035 & -0.005 ± 0.058 & -0.004 ± 0.026 \\
              & 10000     &                   & 0.001 ± 0.004  & 0.001 ± 0.031  & 0.000 ± 0.031  & 0.001 ± 0.035  & 0.001 ± 0.053  & 0.003 ± 0.026  \\
              & 20000     &                   & 0.001 ± 0.003  & -0.002 ± 0.024 & -0.002 ± 0.024 & -0.000 ± 0.030 & -0.004 ± 0.040 & -0.005 ± 0.022 \\
              & 30000     &                   & 0.000 ± 0.002  & 0.000 ± 0.021  & -0.000 ± 0.021 & 0.003 ± 0.027  & 0.004 ± 0.034  & 0.002 ± 0.020  \\
              & 50000     &                   & 0.001 ± 0.002  & 0.001 ± 0.017  & 0.001 ± 0.017  & 0.004 ± 0.025  & 0.004 ± 0.025  & 0.002 ± 0.019  \\
              &           &                   &                &                &                &                &                &                \\
 Total Correlation ($\gamma$)    & 2000      & (inactive→active) & 0.012 ± 0.009  & 0.036 ± 0.019  & 0.034 ± 0.019  & 0.096 ± 0.020  & 0.191 ± 0.024  & 0.072 ± 0.017  \\
              & 5000      &                   & 0.009 ± 0.009  & 0.068 ± 0.027  & 0.067 ± 0.027  & 0.130 ± 0.019  & 0.236 ± 0.024  & 0.096 ± 0.015  \\
              & 10000     &                   & 0.011 ± 0.003  & 0.067 ± 0.027  & 0.067 ± 0.027  & 0.126 ± 0.021  & 0.215 ± 0.023  & 0.095 ± 0.016  \\
              & 20000     &                   & 0.004 ± 0.003  & 0.042 ± 0.022  & 0.042 ± 0.022  & 0.093 ± 0.022  & 0.152 ± 0.022  & 0.068 ± 0.016  \\
              & 30000     &                   & 0.002 ± 0.002  & 0.028 ± 0.020  & 0.028 ± 0.020  & 0.075 ± 0.022  & 0.121 ± 0.021  & 0.058 ± 0.016  \\
              & 50000     &                   & 0.002 ± 0.002  & 0.012 ± 0.017  & 0.012 ± 0.017  & 0.051 ± 0.022  & 0.080 ± 0.018  & 0.042 ± 0.016  \\
              &           &                   &                &                &                &                &                &                \\
 Orthogonality ($\alpha$) & 2000      & (inactive→active) & 0.000 ± 0.009  & 0.013 ± 0.020  & 0.012 ± 0.020  & 0.031 ± 0.028  & 0.058 ± 0.047  & 0.026 ± 0.023  \\
              & 5000      &                   & 0.006 ± 0.009  & 0.021 ± 0.031  & 0.022 ± 0.031  & 0.031 ± 0.034  & 0.055 ± 0.056  & 0.023 ± 0.026  \\
              & 10000     &                   & 0.005 ± 0.004  & 0.019 ± 0.031  & 0.019 ± 0.031  & 0.030 ± 0.034  & 0.050 ± 0.052  & 0.022 ± 0.026  \\
              & 20000     &                   & -0.001 ± 0.003 & 0.016 ± 0.024  & 0.016 ± 0.024  & 0.030 ± 0.029  & 0.049 ± 0.039  & 0.022 ± 0.021  \\
              & 30000     &                   & 0.002 ± 0.002  & 0.006 ± 0.021  & 0.006 ± 0.021  & 0.014 ± 0.027  & 0.028 ± 0.034  & 0.010 ± 0.020  \\
              & 50000     &                   & -0.001 ± 0.002 & 0.003 ± 0.017  & 0.003 ± 0.017  & 0.008 ± 0.025  & 0.013 ± 0.025  & 0.005 ± 0.019  \\
              &           &                   &                &                &                &                &                &                \\
 KL Annealing & 2000      & (inactive→active) & -0.007 ± 0.009 & -0.002 ± 0.020 & -0.003 ± 0.020 & -0.002 ± 0.029 & 0.001 ± 0.049  & -0.006 ± 0.023 \\
              & 5000      &                   & -0.004 ± 0.009 & -0.004 ± 0.031 & -0.005 ± 0.031 & -0.003 ± 0.035 & -0.002 ± 0.058 & -0.002 ± 0.026 \\
              & 10000     &                   & 0.000 ± 0.004  & 0.005 ± 0.031  & 0.005 ± 0.031  & 0.004 ± 0.035  & 0.008 ± 0.053  & 0.005 ± 0.026  \\
              & 20000     &                   & -0.000 ± 0.003 & -0.000 ± 0.024 & -0.000 ± 0.024 & -0.002 ± 0.030 & -0.002 ± 0.040 & 0.003 ± 0.022  \\
              & 30000     &                   & 0.000 ± 0.002  & 0.001 ± 0.021  & 0.001 ± 0.021  & 0.004 ± 0.027  & 0.006 ± 0.034  & -0.001 ± 0.020 \\
              & 50000     &                   & 0.000 ± 0.002  & 0.001 ± 0.017  & 0.001 ± 0.017  & 0.002 ± 0.025  & 0.001 ± 0.025  & -0.000 ± 0.019 \\
              &           &                   &                &                &                &                &                &                \\
\hline
\end{tabular}
\normalsize
\end{table}

\section{Discussion and Conclusion}

This paper introduces SE-VAE, a Structural Equation–Variational Autoencoder that advances white-box generative modeling for tabular data. Grounded in measurement theory, SE-VAE departs from conventional disentanglement approaches by embedding modularity, interpretability, and construct alignment directly into its architecture. Rather than relying heavily on statistical penalties to discover disentangled factors, SE-VAE uses a partitioned encoder-decoder design, a shared nuisance separation mechanism, and global contextual encoding to structurally enforce alignment between observed indicators and their corresponding latent constructs.

Across comprehensive experiments, SE-VAE demonstrates strong and consistent advantages over leading baselines—including $\beta$-VAE, FactorVAE, DIP-VAE, VFAE, and the standard VAE—on six disentanglement metrics. These results hold across a wide range of sample sizes and latent dimensionalities, illustrating the model’s scalability and robustness. While prior methods depend on tuning regularization strengths to navigate entanglement, SE-VAE’s performance emerges from its architectural alignment with known measurement structures. This is further confirmed by ablation experiments, which show that toggling individual loss components (e.g., KL penalty, TC regularization, orthogonality constraint) yields relatively minor performance changes compared to the presence or absence of the architectural structure itself.

The model draws conceptual inspiration from Structural Equation Modeling (SEM), where item–construct relationships are specified in advance. However, while SEM is limited by linearity, fixed loadings, and parametric constraints, SE-VAE generalizes this paradigm to flexible, nonlinear, and scalable inference using deep generative modeling. In doing so, it brings the interpretive clarity of SEM into the modern representation learning toolkit—an advancement particularly suited for fields where latent constructs are theory-driven and validation requires clear correspondence between learned representations and their conceptual definitions.

SE-VAE’s modularity is more than an architectural choice—it is a bridge between domain knowledge and statistical modeling. In many applied domains, researchers conceptualize observed variables as manifestations of latent traits, pathways, or conditions. These groupings are not arbitrary, and models that ignore them risk learning entangled, uninterpretable, or spurious factors. SE-VAE recognized these boundaries by assigning each group of indicators a dedicated sub-encoder and decoder pathway. The inclusion of a shared nuisance latent $z_m$ ensures that global variation—whether due to measurement error, confounding, or method bias—can be modeled explicitly and kept separate from the constructs of interest.

This design is especially impactful in tabular data settings, where interpretability is often a primary goal. In contrast to image or audio domains where disentanglement may be visual or qualitative, tabular settings demand precise alignment with underlying theoretical constructs. SE-VAE meets this demand with a structured architecture that aligns with item design and measurement logic, offering a transparent, reliable foundation for downstream tasks such as latent profiling, fairness analysis, measurement validation, or causal inference.

Beyond its methodological contributions, SE-VAE opens a path for applying deep generative modeling in domains where traditional VAEs have underperformed due to their opacity.  In biomedical research, the model could uncover modular latent pathways from gene expression or clinical indicators, aiding in diagnosis or disease subtyping. In economics or political science, SE-VAE could be applied to behavioral indicators to infer latent ideologies or preferences with interpretability and robustness. In psychology and education, SE-VAE could be used to model latent traits such as motivation, ability, or personality from survey or assessment data, while explicitly accounting for bias or method variance.

SE-VAE is also well-suited to high-stakes domains where transparency and accountability are paramount. Its interpretability allows practitioners to interrogate the meaning of each latent factor, trace its influence on predictions or reconstructions, and validate it against theoretical or empirical expectations. Because it is grounded in structural priors rather than opaque loss optimization, SE-VAE enables domain experts—not just data scientists—to reason about model outputs.

In summary, SE-VAE offers a principled, interpretable, and scalable solution to disentangled representation learning in tabular domains. We hope SE-VAE will serve as a foundation for continued integration of deep learning with scientific and social theory. By uniting the flexibility of neural models with the rigor of measurement logic, SE-VAE contributes not only to the technical advancement of representation learning, but to its alignment with the values of explanation, meaning, and domain relevance.

\bibliography{reference}  
\end{document}